\documentclass[lettersize,journal,preprint]{IEEEtran}
\usepackage{amsmath,amsfonts}
\usepackage{algorithmic}
\usepackage{algorithm}
\usepackage{array}
\usepackage[caption=false,font=normalsize,labelfont=sf,textfont=sf]{subfig}
\usepackage{textcomp}
\usepackage{stfloats}
\usepackage{url}
\usepackage{verbatim}
\usepackage{graphicx}
\usepackage{booktabs}
\usepackage{enumitem}
\usepackage{makecell}
\usepackage{cite}
\usepackage{tikz}
\usetikzlibrary{calc}

\DeclareMathOperator*{\argmax}{argmax}
\begin{document}

\title{Data-Driven Decoding of Russell’s Circumplex Model of Affect}

\author{Amdjed Belaref$^1$ $^2$, Samir Sadok$^3$, Zineb Noumir$^1$, and Renaud Seguier$^2$
\\
    {\normalsize
    $^1$ Alten, France,
    $^2$ CentraleSupélec IETR UMR CNRS 6164,  France\\
    $^3$ Inria at Univ. Grenoble Alpes, CNRS, LJK, France}
\\[1.5ex]
{\normalsize\itshape This work has been submitted to the IEEE for possible publication. Copyright may be transferred without notice, after which this version may no longer be accessible.}
}




\maketitle
\begin{abstract}
Affective computing increasingly relies on deep learning to represent emotions, yet latent spaces often remain opaque, high-dimensional black boxes. This paper investigates whether Transformers' embeddings recover the geometric regularities of Russell’s circumplex model. We unify two complementary experiments testing the hypothesis that, after training models on text and speech, their resulting latent spaces encode a topology consistent with valence-arousal and reproduce human-like neighborhood relations. Specifically, we evaluate deep representations extracted from Transformer-based text (RoBERTa) and speech (wav2vec 2.0) encoders, along with a multimodal Transformer fusion architecture, across naturalistic datasets like MSP-Podcast and controlled LLM-generated stimuli. Our analysis reveals that multimodal fusion of text and audio yields perfect topological alignment with Russell’s primary emotion ordering. Furthermore, in a zero-shot setting using generic text embeddings, projected fine-grained emotion terms fall close to their established human-mapped coordinates. \textbf{Our contribution is a novel, data-driven framework for validating emotion models, demonstrating that Russell's circumplex structure is intrinsically encoded in the embeddings of these modalities rather than being solely an artifact of human labeling, thereby bridging the gap between psychological theory and representation learning.}
\end{abstract}

\section{Introduction}

Understanding human emotion is a complex and nuanced task; Russell's circumplex model of affect provides a theoretical foundation for this by organizing emotions into a circular structure. Using a data-driven approach, our study aims to understand how emotions are organized in the latent spaces of Transformer-based models and whether this organization aligns with Russell’s psychologically based organization.

Advances regarding emotional structures in the field of affective computing have been greatly impacted by various models representing emotions \cite{picard1997affective}, including Plutchik's wheel of emotions \cite{plutchik1980psychoevolutionary}, Ekman's model \cite{ekman1992basic}, and Russell's circumplex model of affect \cite{russell1980circumplex}. Plutchik's model arranges eight primary emotions in pairs of opposites and considers their intensity and combinations. Somewhat similarly, Ekman's model identifies six basic universally expressed and recognized human emotions. In contrast, Russell's model represents emotions in a circular structure where each emotion is positioned according to its valence and arousal level.

Unlike the other emotion representation models, Russell's circumplex model of affect emerged as a data-driven approach, stemming from psychological research on how humans experience and express emotions \cite{russell1980circumplex}. Notably, the model provides empirical evidence that affective concepts can be represented spatially in a \emph{continuous} circumplex, offering a dimensional approach to studying a more comprehensive range of affective states \cite{posner2005a}. In addition, Russell's model offers finer granularity in distinguishing intermediate affective states, providing a more expressive representation than Ekman's model, which explains its widespread use in emotion detection systems.

In the last decade, the circumplex model of affect has been used by many deep learning and machine learning approaches to predict emotions on a dimensional scale \cite{chen2020transformer,schoneveld2021,Hwooi9881519,Wagner2023}. Especially, Transformer-based models, with their high-dimensional embedding spaces, are well-suited to capture such a dimensional representation of emotions \cite{HAZMOUNE2024108339,belaref2024multimodal}. The self-attention mechanism allows for the learning of complex relationships between emotional states. 


Our ambition behind this work is to address the fundamental challenge of Artificial Intelligence explainability through the interpretation of latent spaces. Specifically, we investigate whether it is possible to identify a meaningful manifold within the embeddings produced by Transformer-based models, using the circumplex model as a reference topology. By demonstrating that learned representations naturally align with this theoretical manifold, we aim to open a path for future research where transitions between emotional states could be modeled as consistent transformations in latent space comparable to semantic operations in word embeddings (e.g., country-capital relations). To lay the groundwork for this vision, we investigate whether modern embeddings recover the geometric regularities of Russell’s circumplex through a data-driven reproduction of his original experiments.

A preliminary version of the reproduction of the circular ordering task was presented in \cite{belaref2025circumplex}. The reproduction of the category sort task introduced in section \ref{sec:cat-sort} extends this work by developing a unified analytical framework and providing a comprehensive discussion that bridges the two experiments.

\section{Background and Related Work}

In his 1980 study, Russell conducted experiments that provided key evidence for his proposed circumplex model of affect \cite{russell1980circumplex} and has had a significant impact on affective computing. Criticism surrounding his model is part of an ongoing debate and an area for further research \cite{fontaine2007emotions,Remmingt2000on,LaRowe}. This includes efforts to refine or validate the model or propose alternative frameworks to better understand the structure and relationships among emotions.
One criticism of Russell's circumplex model of affect is its potential oversimplification of emotions \cite{Remmingt2000on}. By reducing the complexity of human emotions to just two dimensions, this model may only partially capture the intricacies of emotional experiences. Other studies like Trnka et al. \cite{trnka2016a} attempt to refine this model by introducing a 3D hypercube projection. This adds two more dimensions to provide a more comprehensive representation that captures the complexity of emotion more accurately. In Fontaine's study \cite{Johnny2022r} of the linear and nonlinear relationship of the cognitive-emotional structure, the authors suggest that while a two-dimensional structure fits emotional data well, a four-dimensional representation is needed for a comprehensive, full-range representation. While these higher-dimensional models capture greater emotional nuance, they lack the strict geometric regularities required for our validation approach.

We are aware that current emotion representation models may be limited by their simplistic modeling of emotions, failing to address all the nuances and ambiguities of emotional experiences \cite{tran2022h,izard2008emotion}.  Sethu et al. \cite{Sethu2019TheAW} propose a framework to address this issue by introducing an approach designed to describe and reason about emotional representations. However, for the purpose of validating learned representations, the circumplex remains the most robust \emph{minimal topological constraint}. Its circularity imposes specific neighborhood and opposition rules (e.g., excitement vs. depression) that serve as a rigorous test for the coherence of learned embeddings.

Recent work by Wagner et al. \cite{Wagner2023} highlights that Transformer-based models have significantly closed the valence gap in speech emotion recognition, suggesting that these models implicitly learn sophisticated affective representations. Yet, these latent spaces often remain opaque, high-dimensional black boxes. Another relevant study showed, through the visualization of speech features on a low-dimensional manifold, consistency with Russell's dimensional model of emotion \cite{Vayryn2013en}. Our work extends this inquiry by interrogating whether modern Self-Supervised Learning (SSL) models spontaneously recover this specific psychological geometry without explicit supervision. Our research question extends beyond automation: \emph{Does a coherent affective manifold emerge within the Transformer’s latent space that we can exploit to drive interpretability?}

Instead of relying on human feedback to construct the circumplex, we investigate whether the intrinsic geometry of the model's final layers spontaneously recovers the circular topology proposed by Russell. Through a data-driven deep learning approach, we test whether this latent structure allows us to reproduce Russell’s tasks without human intervention, thereby validating that the psychological dimensions of affect are naturally encoded in high-dimensional representation spaces.



\section{Russell's Original Experiments}
\label{sec:russell_experiments}
Russell's experiments consist of two tasks that are shaped by the students' perception of emotion and are influenced by their lifelong knowledge and prior experiences. They first sorted 28 mood terms into eight predefined primary emotions in the \emph{category sort task}; then they arranged these eight categories around a circle so that opposites lay across from each other and similar terms lay adjacent in the \emph{circular ordering task} \cite{russell1980circumplex}. This makes his model particularly interesting because it taps into an innate aspect of human nature, yielding an emotional circular ordering. We summarize both protocols below, as they serve as the reference targets for the data-driven reproductions
presented in Section~\ref{sec:methodology}.

\subsection{Circular Ordering Task}
\label{sec:russell_circular}
This ordering task stated the following for the 36 undergraduates who participated in the study: start with \emph{Aroused} in the first position, and after that, arrange the remaining affective words (excited, pleased, contented, sleepy, depressed, miserable, distressed) by meeting two criteria, such that words opposite each other on the circle describe opposite feelings, while words closer together on the circle describe more similar feelings. The question of how to apply this process in a data-driven context addresses whether Russell’s circumplex model reflects an inherent structure of emotional experience or one imposed by theory.

\begin{figure}[htpb!]
  \centering
  \includegraphics[width=0.85\linewidth]{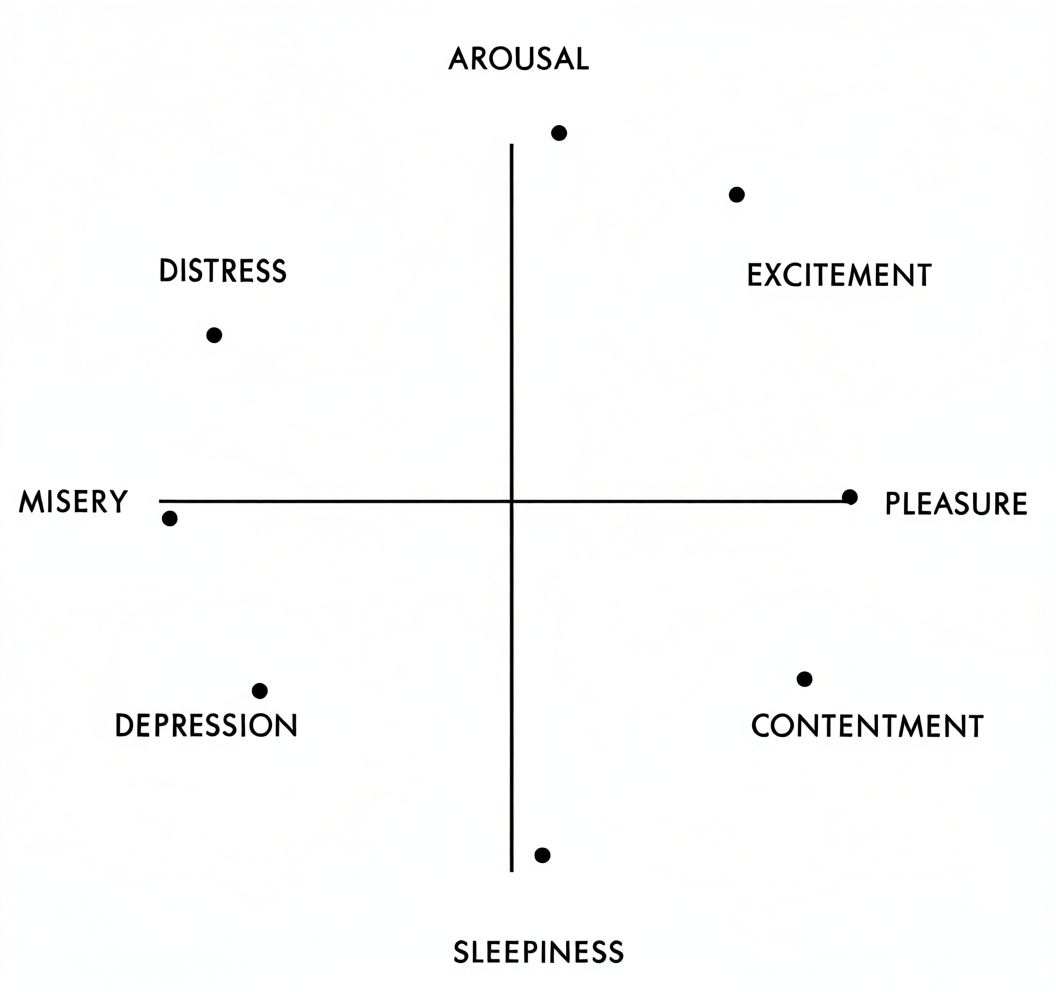}
  \caption{Circular order task results of Russell's Eight affect concepts.}
  \label{fig:russells_arrangement}
\end{figure}

\begin{table}[h]
    \centering
    \caption{Russell's result table: Frequency With Which Emotion Terms Were Placed in Eight Positions Around a Circle}
    \label{tab:russell-results}
    \begin{tabular}{|l|c|c|c|c|c|c|c|c|}
    \hline
    \textbf{Term} & \textbf{1} & \textbf{2} & \textbf{3} & \textbf{4} & \textbf{5} & \textbf{6} & \textbf{7} & \textbf{8} \\
    \hline
    Aroused & \textbf{36} &  &  &  &  &  &  &   \\
    Excited &  & \textbf{24} & 3 & 1 &  &  &  & 8 \\
    Pleased &  & 9 & \textbf{20} & 7 &  &  &  &  \\
    Contented &  & 2 & 13 & \textbf{16} & 3 &  &  & 2 \\
    Sleepy &  &  &  & 9 & \textbf{23} & 3 &  & 1 \\
    Depressed &  & 1 &  &  & 5 & \textbf{19} & 10 & 1 \\
    Miserable &  &  &  & 1 & 1 & 11 & \textbf{18} & 5  \\
    Distressed &  &  &  & 2 & 4 & 3 & 8 & \textbf{19} \\
    \hline
    \end{tabular}; 

\end{table}

Table~\ref{tab:russell-results} details the frequency with which participants assigned emotion terms to the eight circular positions. Russell's theoretical order (\textbf{Aroused, Excited, Pleased, Contented, Sleepy, Depressed, Miserable, Distressed}) is mapped to the numerical position sequence \textbf{1/2/3/4/5/6/7/8}. Figure~\ref{fig:russells_arrangement} shows the resulting circular arrangement of Russell's eight affect concepts.

\subsection{Category Sort Task}
\label{sec:russell_category}
In Russell's Category Sort Task, we distinguish two levels of emotion granularity; the primary categories are the eight
circumplex affective states: \(\mathcal{C}\) = \{Aroused, Excited, Pleased, Contented, Sleepy, Depressed, Miserable, Distressed\}. In contrast, fine-grained terms are 28 specific emotion words (e.g., \emph{Happy, Delighted, Frustrated}), capturing subtle variations within the broader affect space. In the experimental protocol, the 36 participants were asked to categorize the nuanced emotion terms into the primary category that best fit each one. This sorting frequency allowed Russell to observe how people naturally group emotions, revealing how they perceive semantic proximities within the affective space.

\begin{figure}[h]
  \centering
  \includegraphics[width=\linewidth]{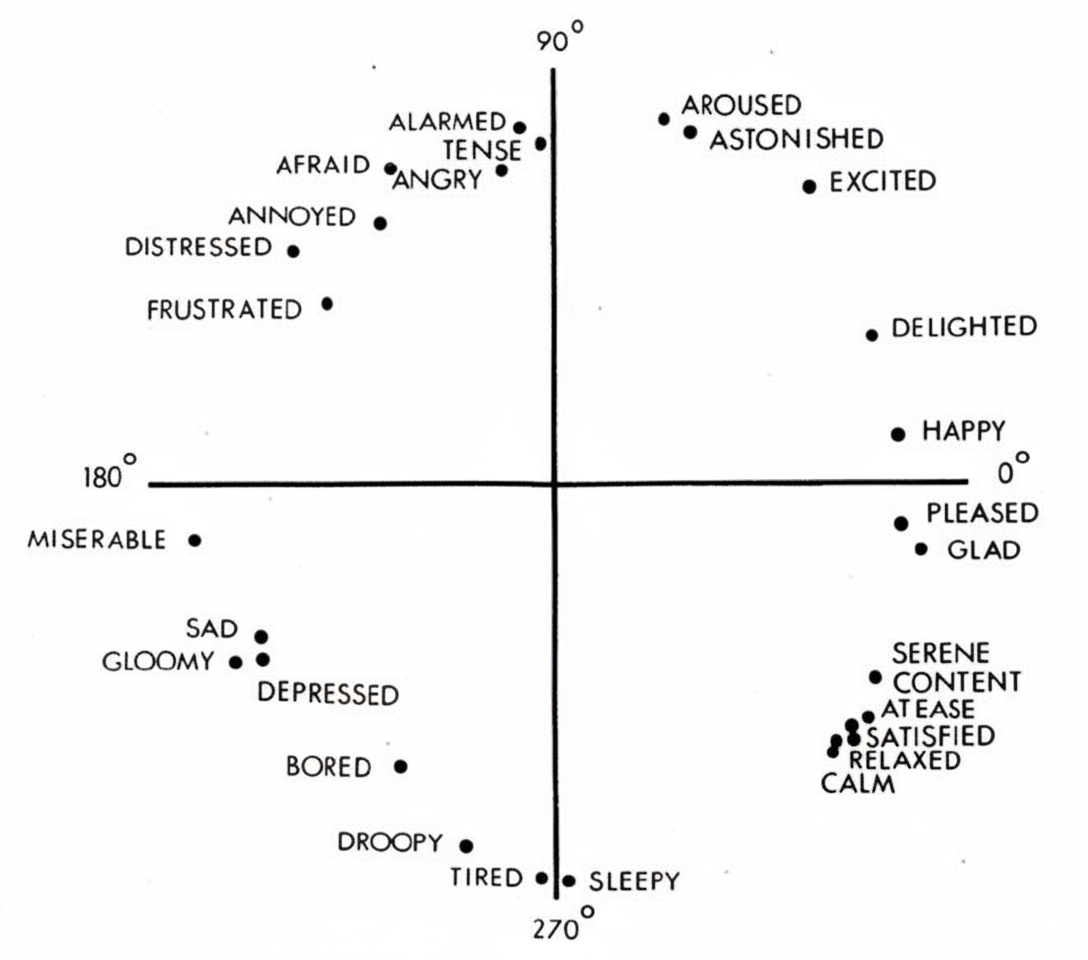}
  \caption{Direct circular scaling coordinates for 28 affect words \cite{russell1980circumplex}}
  \label{fig:28emotion}
\end{figure}

To generate the graphical representation in Figure~\ref{fig:28emotion}, Russell first collected sorting data. He aggregated these manual classifications into a frequency table (see Table~\ref{tab:freq_28_words_compact}) and applied a circular statistical transformation~\cite{ross1938circular}. This method converted the discrete assignment probabilities into polar coordinates, defining the precise angular location and concentration (precision) of each term on the circumplex.

\begin{table}[htbp]
\centering
\caption{Frequency of Placement of 28 Words Into Eight Categories \cite{russell1980circumplex}}
\label{tab:freq_28_words_compact}

\begingroup
\scriptsize                              
\setlength{\tabcolsep}{2pt}              
\renewcommand{\arraystretch}{0.82}       

\begin{tabular}{@{}l*{8}{c}@{}}          
\toprule
\multicolumn{1}{c}{Term} & \multicolumn{8}{c}{Category}\\
\cmidrule(lr){2-9}
 & Pleasure & Excitement & Arousal & Distress & Misery & Depression & \makecell{Sleepi-\\ness} & \makecell{Content-\\ment}\\
\midrule
Happy      & 21 &  8 &  2 &    &    &    &    &  5 \\
Delighted  & 15 & 16 &  3 &    &    &    &    &  2 \\
Excited    &  2 & 29 &  5 &    &    &    &    &    \\
Astonished &    & 17 & 18 &  1 &    &    &    &    \\
Aroused    &    & 14 & 21 &  1 &    &    &    &    \\
Tense      &    &  8 & 18 &  9 &    &  1 &    &    \\
Alarmed    &    &  6 & 19 & 11 &    &    &    &    \\
Angry      &    &  5 & 21 &  5 &  3 &  2 &    &    \\
Afraid     &    &  2 & 11 & 22 &    &  1 &    &    \\
Annoyed    &    &  1 & 12 & 14 &  4 &  4 &    &  1 \\
Distressed &    &    &  4 & 25 &  5 &  2 &    &    \\
Frustrated &    &  2 &  5 & 19 &  4 &  6 &    &    \\
Miserable  &    &    &    &  3 & 23 & 10 &    &    \\
Sad        &    &    &    & 10 &  6 & 19 &    &  1  \\
Gloomy     &    &    &    &  2 & 11 & 22 &  1 &    \\
Depressed  &    &    &    &  4 &  7 & 24 &    &  1 \\
Bored      &    &    &    &  3 &  2 & 14 & 17 &    \\
Droopy     &    &    &    &  1 &  1 &  8 & 26 &    \\
Tired      &    &    &    &    &  1 &  1 & 34 &    \\
Sleepy     &    &    &    &    &  1 &    & 32 &  3 \\
Calm       &  4 &    &    &    &    &    & 3  & 29 \\
Relaxed    &  6 &    &    &    &    &    & 4  & 26 \\
Satisfied  &  3 &  1 &    &    &    &    &    & 32 \\
At ease    &  7 &    &    &    &    &    & 3  & 26 \\
Content    &  6 &  1 &    &    &    &    &    & 29 \\
Serene     &  8 &  2 &    &    &    &    &    & 26 \\
Glad       & 20 &  4 &    &    &    &    &    & 12 \\
Pleased    & 22 &  2 &  2 &    &    &    &    & 10 \\
\bottomrule
\end{tabular}
\endgroup
\end{table}

It is important to understand that, in this table, Russell's objective was not merely descriptive but aimed at revealing relational structure, demonstrating that human affective sensitivity follows a continuous geometric pattern.

Conceptually, this mirrors a broader idea in sensory neuroscience: organized maps can be inferred from systematic response patterns (as in Hubel and Wiesel~\cite{hubel1959receptive} demonstrations of orderly feature preferences in the visual cortex). Here, the frequency table plays a similar role as an empirical mapping of neighboring relations among affective categories.

In the following sections, we present our data-driven frameworks for reproducing both tasks using Transformer-based embeddings.

\section{Framework 1: Circular Ordering Task}
\label{sec:methodology}
\begin{figure*}[t!]
  \centering
  \includegraphics[width=1\linewidth]{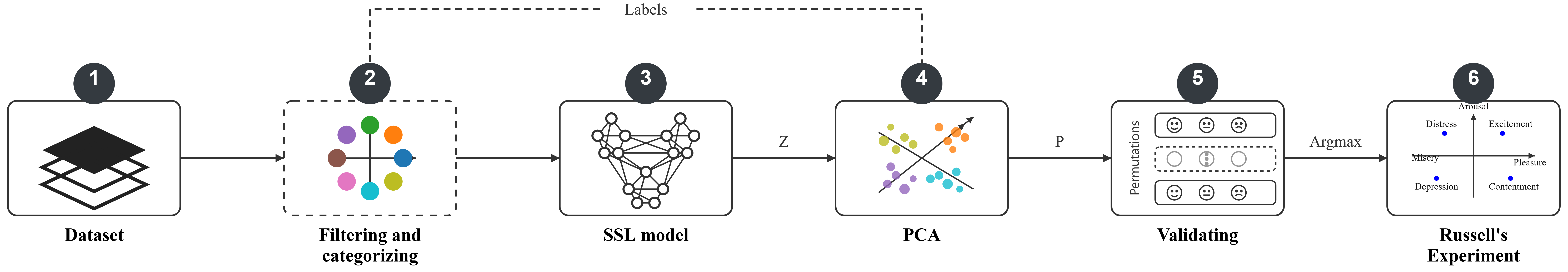}
  \caption{Reproduction framework of the Circular Ordering Task }
  \label{fig:approach}
\end{figure*}

We reproduce the circular ordering task described in Section~\ref{sec:russell_circular} in both unimodal and multimodal settings, working with textual and acoustic data. Figure \ref{fig:approach} illustrates the overview of our approach, highlighting the key transformation stages that enable our data-driven reproduction. The pipeline begins with a dataset containing emotional expressions and concludes with a validated circular arrangement of the eight primary emotions.

\subsection{Methodology of Framework 1}
\label{sec:circular_task}


\noindent\textbf{Datasets.}\hspace{0.3cm} Existing annotated emotion datasets exhibit significant diversity and unique characteristics that impact the generalizability and applicability of the results. These variations include whether the emotions are spontaneous or acted, whether they are captured in controlled or naturalistic settings, the modalities recorded, the annotation methods employed, and the range of emotions represented. As stated above, we are working with textual \emph{and} acoustic data for our experiments. For our study, we evaluated three datasets defined as follows:

\begin{itemize}[leftmargin=*]
    \item \textit{CoLiTec corpora (text-based)} \cite{FerraresiBernardini2013}: A large web-derived corpus, partially publicly available, from which we extracted a dataset of sentences. The phrases were centered around the eight affective words found on Russell's model $y_n =$ \{aroused, excited, pleased, contented, sleepy, depressed, miserable, distressed\}, with 50 sentences for each word.
    \item \textit{TESS (audio-based)} \cite{pichora2020tess}: A dataset of 200 target words spoken in the carrier phrase "Say the word $\texttt{<\dots>}$" by two actresses (aged 26 and 64 years). The recordings illustrated these emotions $y_n =$ \{anger, disgust, fear, happiness, pleasant surprise, sadness, and neutral\}. There are 2800 stimuli in total with categorical annotation.
    \item \textit{MSP-Podcast (text and audio)} \cite{lotfian2017msp}: The largest naturalistic speech emotional corpus contains speech segments extracted from podcast recordings. The corpus includes transcriptions and comprises 151,654 speaking turns, totaling 237 hours and 56 minutes. The data is labeled with both categorical and dimensional emotion representations. Given that the MSP-Podcast dataset is large and has unbalanced emotion classes, we decided to create a subset of this dataset. This subset is balanced between emotion classes $y_n =$ \{happy, angry, sad, neutral\} and includes labeled data with a consensus percentage above 50\%. After creating this subset, the final number of speaking turns was reduced to 9,673.
\end{itemize}



\begin{figure}[t!] 
  \centering
  \includegraphics[width=\linewidth]{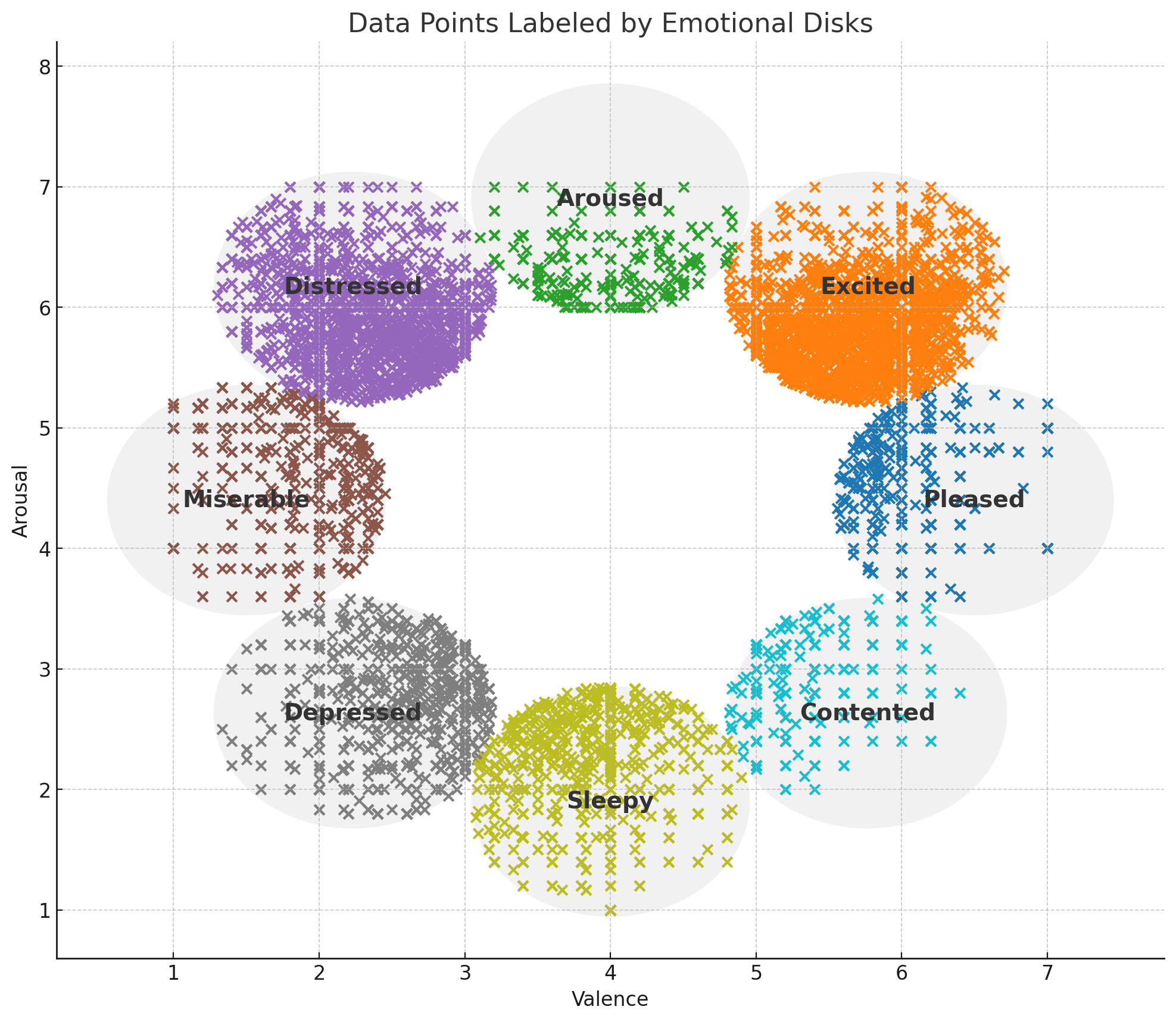} 
  \caption{This figure illustrates the applied Russell-based filter.}
  \label{fig:russel-filter}
\end{figure}

\noindent\textbf{Filtering and Categorizing Following Russell Representation.}\hspace{0.2cm}
\label{sec:step-1}
Before turning to the representation learning phase, we must align the data labels with Russell's eight primary affective words. The data used must have continuous valence and arousal scores to employ a filtering approach that maps these continuous coordinates onto Russell's categories. 
We operationalize Russell’s model by defining eight fixed \emph{emotional centers} in the Valence–Arousal (V–A) plane, each placed at the theorized coordinates of a primary emotion (e.g., Pleasure at 0°, Excitement at 45°, Arousal at 90°). To categorize the data, we define an acceptance region: a circular disk of fixed radius centered at each emotional center, as illustrated in Figure \ref{fig:russel-filter}. We project each data point into the Valence--Arousal (V--A) plane using its annotated valence and arousal scores, then filter and label points as follows: Each sample is first mapped to a two-dimensional coordinate $(v, a)$. 
A spatial filtering step then determines whether the point lies within a fixed-radius disk centered around any of the eight emotional anchors defined in Russell’s circumplex model. 
If the coordinate falls inside one of these disks, the corresponding Russell category is assigned (e.g., points inside the $0^\circ$ disk are labeled as \emph{Pleased}). 
Samples located outside all predefined disks—corresponding to ambiguous regions of the affective space—are discarded.
These assigned labels allow us to filter out the data that will not be used and serve strictly as analytical tags to group embeddings for centroid calculation and to color-code the latent space for visualization (as seen in Figure \ref{fig:russel-filter}).\\

\noindent\textbf{Deep Representation Learning.}\hspace{0.2cm} We encode each sample $x_n$ with a deep model to obtain its learned representation $\mathbf{z}_n$. This representation is extracted using Transformer-based models \cite{liu2019roberta,baevski2020wav2vec}. They have shown significant promise in extracting and fusing features from textual and acoustic modalities for tasks such as emotion recognition \cite{HAZMOUNE2024108339,belaref2024multimodal} .
In our study, we employed RoBERTa \cite{liu2019roberta} and wav2vec \cite{baevski2020wav2vec} for text and audio data, respectively. We use these architectures in two distinct configurations: generic (pre-trained only on self-supervised objectives) and fine-tuned (adapted for emotion classification on datasets like RAVDESS, IEMOCAP, TweetEval, and MSP-Podcast). It is important to note that while we utilize models fine-tuned on emotion recognition, we employ them strictly as fixed feature extractors without further training.

\begin{figure}[t!]
  \centering
  \includegraphics[width=0.8\linewidth]{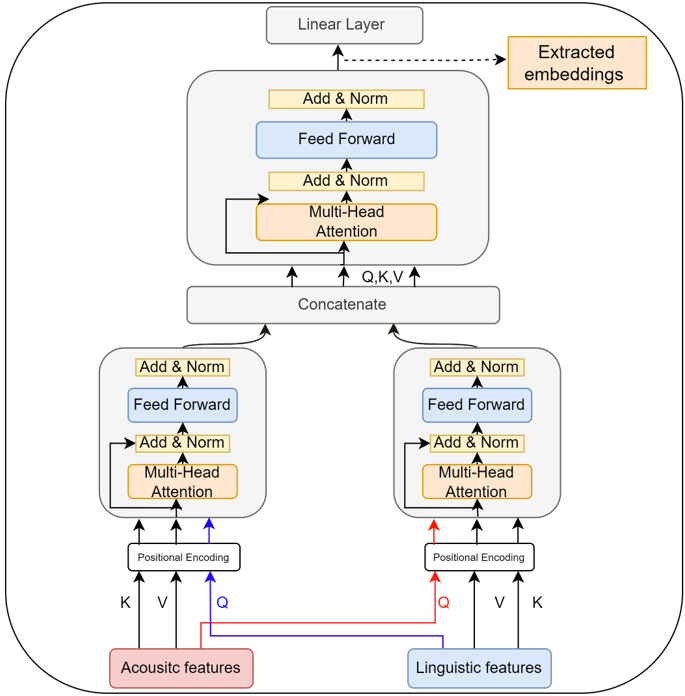}
  \caption{This figure illustrates our Multimodal Transformer fusion architecture and at what level the embedding extraction was done. }
  \label{fig:multimodal-architecture}
\end{figure}

In a multimodal setting, we developed a Transformer fusion architecture using cross-modal attention, inspired by MulT \cite{tsai2019multimodal}. As shown in Figure~\ref{fig:multimodal-architecture}, this approach consists of projecting both audio and text features into a \emph{shared} latent space. We train the inter- and intra-connections between the two modalities using cross-modal attention. The outputs of the cross-modal attention are concatenated and passed through an additional Transformer encoder for further refinement. This process retrieves the fused features, forming a comprehensive representation space for the multimodal data.
It is also important to note that the multimodal phase required a supervised emotion classification training step to build a coherent fused latent space. Because linguistic and acoustic embeddings come from different geometric manifolds, they cannot be easily combined. However, the embeddings we analyze are extracted before the classification head, ensuring that the dimensional structure we validate is not an artifact of the categorical decision boundary but reflects the geometry of the shared representation itself. \\

\noindent\textbf{Dimensionality Reduction.}\hspace{0.2cm} We apply dimensionality reduction techniques to analyze, visualize, and interpret the latent space $\mathbf{z}_n$. We use Principal Component Analysis (PCA), a linear transformation technique that reduces dimensionality while retaining the important information from the original data. It identifies the directions (i.e., principal components) along which the data has the highest variance. This can be interpreted as finding a latent subspace ($\mathbf{p}$) of a very small dimension compared to ($\mathbf{z}$). Specifically, we established the dimension of $\mathbf{p}$ at 10, as empirical analysis demonstrated that the first 10 principal components retain the majority of the data variance. We chose to use PCA over other dimensionality reduction methods due to its linear mapping, ability to preserve global structure, and interpretability.\\

\noindent\textbf{Validating Results with Russell's Model.}\hspace{0.2cm}
We investigate whether the learned emotional representations follow the circular structure proposed in Russell’s circumplex model. To this end, we first compute centered subspace vectors $\mathbf{p}^{(center)}$ for each emotion, which serve as nodes in a latent affective space. We then evaluate whether these nodes admit a circular ordering consistent with Russell’s theoretical arrangement. To do so, we enumerate all possible permutations of the emotional vectors while fixing $\mathbf{p}_{aroused}^{(center)}$ as an anchor point, in accordance with Russell’s reference orientation. The remaining seven emotions (\emph{Pleased} to \emph{Distressed}) are permuted exhaustively, yielding $N = 7! = 5040$ candidate circular sequences. For each permutation $i$, we compute a smoothness score $S_i$ (Eq.~\ref{eq:metric-s}), defined as the sum of cosine similarities between all adjacent emotion pairs along the candidate circle, including the closing link back to \emph{Aroused}. This metric quantifies how smoothly emotions transition from one to the next, under the assumption that neighboring emotions in Russell’s model should exhibit high representational similarity. The optimal latent ordering is obtained by selecting the permutation $i^*$ that maximizes the total smoothness score:
\begin{equation}
    \label{eq:argmax}
    i^* = \argmax_i S_i, \quad \forall i \in \{1, 2, \dots, N\}.
\end{equation}
The smoothness score for permutation $i$ is computed as:
\begin{align}
    \label{eq:metric-s}
    S_i &= \texttt{cos-sim}(\mathbf{p}_{aroused}^{(center)}, \texttt{Per}(1)_i) 
    \nonumber\\
    &+ \sum_{j=1}^{7} \texttt{cos-sim}(\texttt{Per}(j)_i, \texttt{Per}(j+1)_i)  
    \nonumber\\
    &+ \texttt{cos-sim}(\texttt{Per}(7)_i, \mathbf{p}_{aroused}^{(center)}).
\end{align}
Finally, we compare the optimal sequence $i^*$ with Russell’s theoretical circular order and quantify alignment by counting the number of positional mismatches (inversions), thereby measuring the degree to which the learned representation conforms to psychological theory.

\subsection{Results of Framework 1}

\begin{table}[h]
    \centering
    \caption{Russell's Frequency Table with Permissible Inversions Highlighted}
    \label{tab:H-russell-results}
    \begin{tikzpicture}
    \node (table) {
    \begin{tabular}{|l|c|c|c|c|c|c|c|c|}
    \hline
    \textbf{Term} & \textbf{1} & \textbf{2} & \textbf{3} & \textbf{4} & \textbf{5} & \textbf{6} & \textbf{7} & \textbf{8} \\
    \hline
    Aroused & \textbf{36} &  &  &  &  &  &  &   \\
    Excited &  & \textbf{24} & 3 & 1 &  &  &  & 8 \\
    Pleased &  & 9 & \textbf{20} & 7 &  &  &  &  \\
    Contented &  & 2 & 13 & \textbf{16} & 3 &  &  & 2 \\
    Sleepy &  &  &  & 9 & \textbf{23} & 3 &  & 1 \\
    Depressed &  & 1 &  &  & 5 & \textbf{19} & 10 & 1 \\
    Miserable &  &  &  & 1 & 1 & 11 & \textbf{18} & 5  \\
    Distressed &  &  &  & 2 & 4 & 3 & 8 & \textbf{19} \\
    \hline
    \end{tabular}
    };

    \draw[red, thick] ($(table)+(-0.45,0.50)$) rectangle ($(table)+(0.65,-0.10)$);
    \draw[red, thick] ($(table)+(2.8,-1.1)$) rectangle ($(table)+(1.7,-0.48)$);    
    \end{tikzpicture}
\end{table}

The data in Table~\ref{tab:H-russell-results} reveal a clear consensus, with some neighboring emotions appearing in inverted positions; these inversions are explained by humans' different perceptions of close emotional states.
For example, column 3 shows the degree of consensus on \textbf{Pleased}: 20/36 placed it at position 3, while 3 chose \textbf{Excited} and 13 \textbf{Contented}.

To account for this natural variance, we establish four valid reference arrangements for validation. First, two primary circular sequences:

\begin{itemize}
    \item 1/2/3/4/5/6/7/8 (the original sequence)
    \item  1/8/7/6/5/4/3/2 (the reverse sequence)
\end{itemize}

And two permissible variations, accounting for emotions perceived as semantically adjacent (highlighted in red boxes):

\begin{itemize}
    \item  1/2/\textbf{4}/\textbf{3}/5/6/7/8 (sequence with inverted positions of ‘Pleased’ and ‘Contented’)
    \item 1/2/3/4/5/\textbf{7}/\textbf{6}/8 (sequence with inverted positions of ‘Depressed’ and ‘Miserable’ )
\end{itemize}

To validate our approach, we compare the single permutation $i^*$ that maximizes the total similarity score (as defined in Eq.\ref{eq:argmax}) against these four arrangements. We quantify this alignment by calculating the \textit{cyclic mismatch count}, to gain insights on how well our approach aligns with human perceptions. This comparison would validate our method and account for individual differences in emotional perception seen in the original experiment.

Having detailed our validation framework, we now report the results of the circular ordering task across the three selected datasets, evaluating their alignment with Russell’s theoretical model.

\subsubsection{Results on CoLiTec Corpora Text-Extracted Dataset } In unimodal mode, feature extraction was done through a generic BERT model followed by dimensionality reduction. By applying our approach, the configuration yielding the highest similarity score was identified as 1/2/3/4/5/\textbf{7}/\textbf{6}/8, with only one inversion between positions 6 and 7, corresponding to the affective words \textbf{Depressed} and \textbf{Miserable}. Their semantic and emotional similarity, along with the simplicity of the data, in which the exact affective word is included in the phrases, justifies this inversion. The exact inversion exists in Russell's results in Table \ref{tab:russell-results}, which reflects an individual's subjective interpretation or a subtle difference in the intensity or nuances of the emotion.

\subsubsection{Results on TESS } 

In unimodal mode, using a generic wav2vec model on the TESS audio dataset resulted in a mixed, random space with inseparable clusters. However, a fine-tuned wav2vec model on emotion produced an interpretable embedding space with distinct emotion clusters. Although the absence of valence-arousal scores for this dataset prevented the application of the Russell-based filter (Sec.\ref{sec:step-1}), we applied the permutation method (Eq. \ref{eq:metric-s}) directly to the class centroids. The configuration producing the \textbf{highest similarity score} was identified as \textbf{Fear, Angry, Happy, Pleasantly Surprised, Disgusted, Neutral, and Sad}.

Notably, this order reflects a clear decrease in arousal. It starts with high-energy emotions (Fear, Anger, Happy) and ends with low-energy ones (Neutral, Sad). Our hypothesis is that the learned representation prioritizes \textit{activation} as the primary organizing dimension. As if the circular structure were flattened onto the arousal axis.

\subsubsection{Results on MSP-Podcast }In unimodal mode on MSP-Podcast text and audio data:
\paragraph{Audio}: Using a fine-tuned wav2vec model on emotion and applying the Russell-based filter, 1/2/3/4/5/6/7/8 is the best permutation with the highest cosine similarity. This arrangement has no mismatched positions compared to Russell's arrangement.
\paragraph{Text}: Using a fine-tuned RoBERTa model on emotion, the best permutation with the highest cosine similarity is 1/\textbf{4/2/3}/5/6/7/8. This arrangement has three mismatched positions compared to Russell's arrangement.
\begin{figure}[h]
  \centering
  \includegraphics[width=0.9\linewidth]{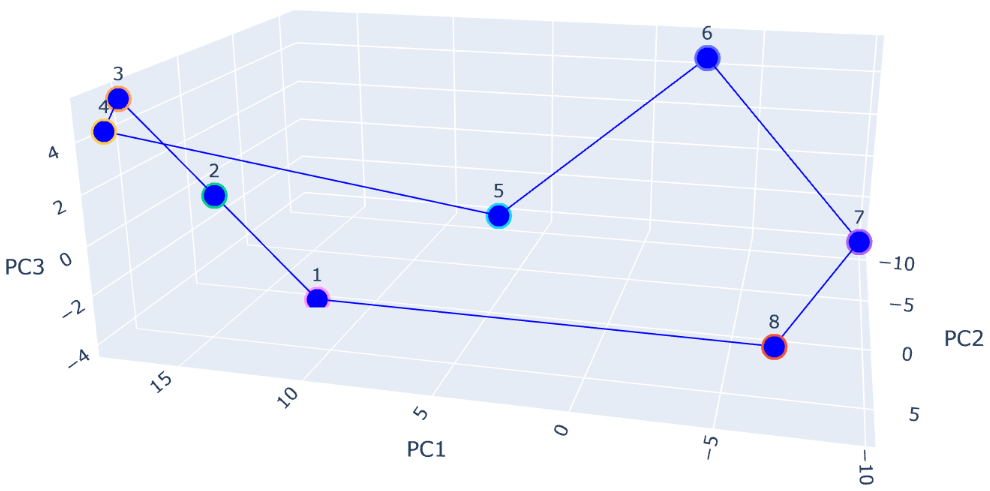}
  \caption{Arrangement of the centered subspace vectors in 3D PCA space with our fusion approach.}
  \label{fig:arrangement}
\end{figure}
\paragraph{Fusion}: Combining features from both modalities with our multimodal architecture results in the emotion centers and arrangement shown in Figure~\ref{fig:arrangement}. The best permutation with the highest similarity score is 1/2/3/4/5/6/7/8. This arrangement has no mismatched positions compared to Russell's arrangement.

\section{Framework 2: Category Sort Task}
\label{sec:cat-sort}
\begin{figure*}[h]
  \centering
  \includegraphics[width=\linewidth]{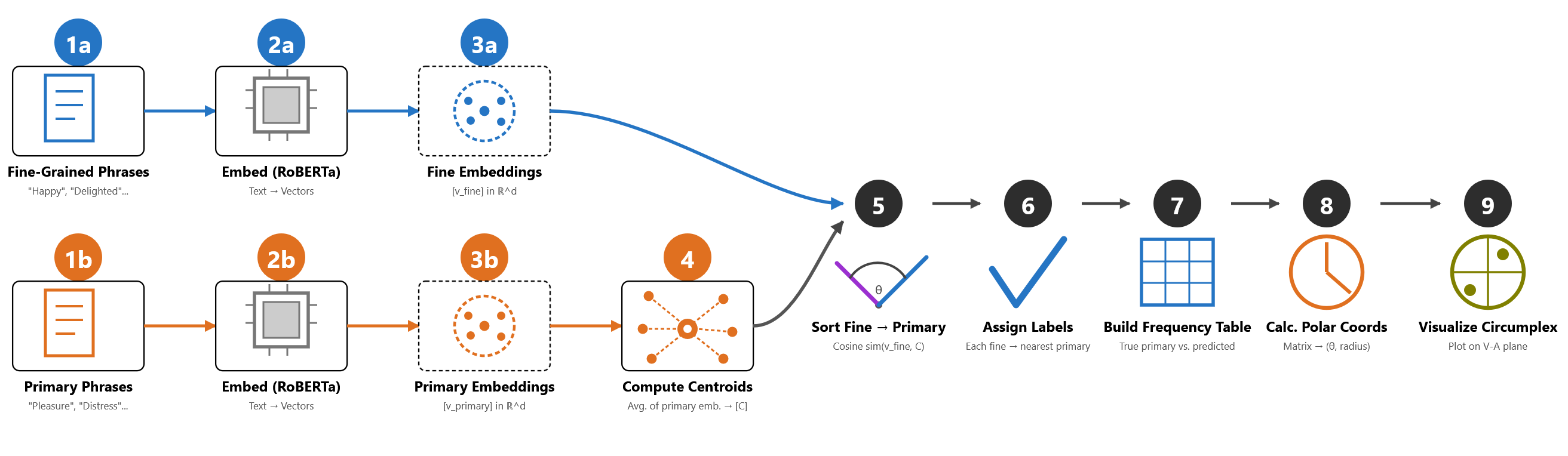}
  \caption{Pipeline of our Data-Driven Reproduction of Russell's Category sort task }
  \label{fig:Pipe}
\end{figure*}

We now reproduce the category sort task (Section~\ref{sec:russell_category}). We translate this psychological experiment into a computational procedure suitable for text embeddings. While the objective remains the same, the mapping of fine-grained terms to primary categories. We adapt the mechanism by replacing human participants' sorting with language model embeddings, along with geometric measurements: given sentence embeddings of fine-grained affective terms, we assign each term embedding to one of the eight primary emotion centroids using cosine similarity. We then analyze the resulting circumplex geometry and compare it to Russell’s original positions. To get an overview of our framework, we summarize our complete pipeline in Figure \ref{fig:Pipe}, which illustrates the end-to-end workflow of our data-driven approach.

\subsection{Method of Framework 2}
\noindent\textbf{Dataset Construction.}\hspace{0.2cm}
In our effort to reproduce the Russell category task, we had to find a way to mirror the sorting of the 28 fine-grained emotion terms in a data-driven manner. The 36 students who participated in the experiment represent, in a sense, 36 different views and perspectives on these 28 emotional terms. To emulate this diversity in a computational context, we decided to have a phrase that conveys one of these nuanced emotions serve as an equivalent to a person's view on it. Thus, we would build a base of 50 phrases per fine-grained emotion to replace the participants' perceptions of them.

\paragraph{LLM-Based Phrase Generation} Since no public corpus contains the specific 28 terms Russell used with sufficient phrase diversity, we constructed a controlled dataset using large language model generation. We employed Anthropic's Claude Sonnet 3.7 \cite{anthropic_claude37_systemcard_2025}, with carefully crafted prompts designed to elicit natural, varied expressions of each emotion while ensuring each phrase remained strictly anchored to the specific definition of the target term.

\noindent\textbf{Representation  and  Assignment.}\hspace{0.2cm}
Since our work is based on the latent representation of a Transformer encoder, the primary emotion categories to which the phrases of 28 terms are to be assigned should be in the same latent space, and they should be represented as an \emph{anchor}. We represent them in our experiment as centroids. For each primary emotion, we curated a set of five phrases and averaged their embeddings to form the centroid of that emotion, making similarity comparisons in the latent space meaningful.


This process involves two distinct steps of aggregation: first, computing a single embedding for each phrase, and second, calculating the center for each emotion category.

\subsubsection{From Tokens to Sentence Embeddings}

We utilize the RoBERTa-base encoder $\phi(\cdot)$. For a given input phrase $s$, let $L$ be the sequence length (number of tokens or words). The model outputs a sequence of vectors from the \textbf{last hidden layer}, denoted as $\mathbf{H} = [\mathbf{h}_1, \dots, \mathbf{h}_L]$. Each vector $\mathbf{h}_i$ exists in a $d$-dimensional space, where $d=768$ (the standard dimension for RoBERTa-base).

To obtain a single representation $\mathbf{e}(s) \in \mathbb{R}^{768}$ for the entire phrase, we calculate the arithmetic mean of the token vectors $\mathbf{h}_i$, weighted by the attention mask $m_i \in \{0,1\}$ (which serves to exclude padding tokens). Unlike the reproduction of the circular ordering task, no dimensionality reduction is applied; we operate directly in the high-dimensional latent space:

\begin{equation}
\mathbf{e}(s) \;=\; \frac{\sum_{i=1}^{L} m_i\, \mathbf{h}_i(s)}{\sum_{i=1}^{L} m_i} \quad \in \mathbb{R}^{768}.
\end{equation}

This step is also applied to extract sentence embeddings for fine-grained phrases. 

\subsubsection{From Sentence Embeddings to Emotion Centroids}

Once every phrase is mapped to a vector in $\mathbb{R}^{768}$, we compute the \emph{emotion centroid} $\boldsymbol{\mu}_c$. This vector represents the geometric center of all phrases belonging to a primary emotion category $c$. For a set $\mathcal{S}_c$ containing $N$ distinct phrases (here $N=5$), the centroid is simply the average of their embeddings:
\begin{equation}
\boldsymbol{\mu}_c \;=\; \frac{1}{\lvert \mathcal{S}_c \rvert}\sum_{s\in\mathcal{S}_c} \mathbf{e}(s) \quad \in \mathbb{R}^{768}.
\end{equation}

Thus, both the fine-term phrase embeddings $\mathbf{e}(s)$ and the category centroids $\boldsymbol{\mu}_c$ exist in the same $768$-dimensional semantic space, allowing for direct cosine similarity comparison.

\subsubsection{Cosine-Similarity Sorting (Assignment Rule)}
We also had to find a way to formulate the sorting done in the original experiment. We developed a mechanism of sorting based on the cosine similarities between the phrase embeddings of a fine-grained emotion  $e(t)$ and the centroids of primary emotion $\mu_c$.

\begin{figure}[h]
  \centering
  \includegraphics[width=\linewidth]{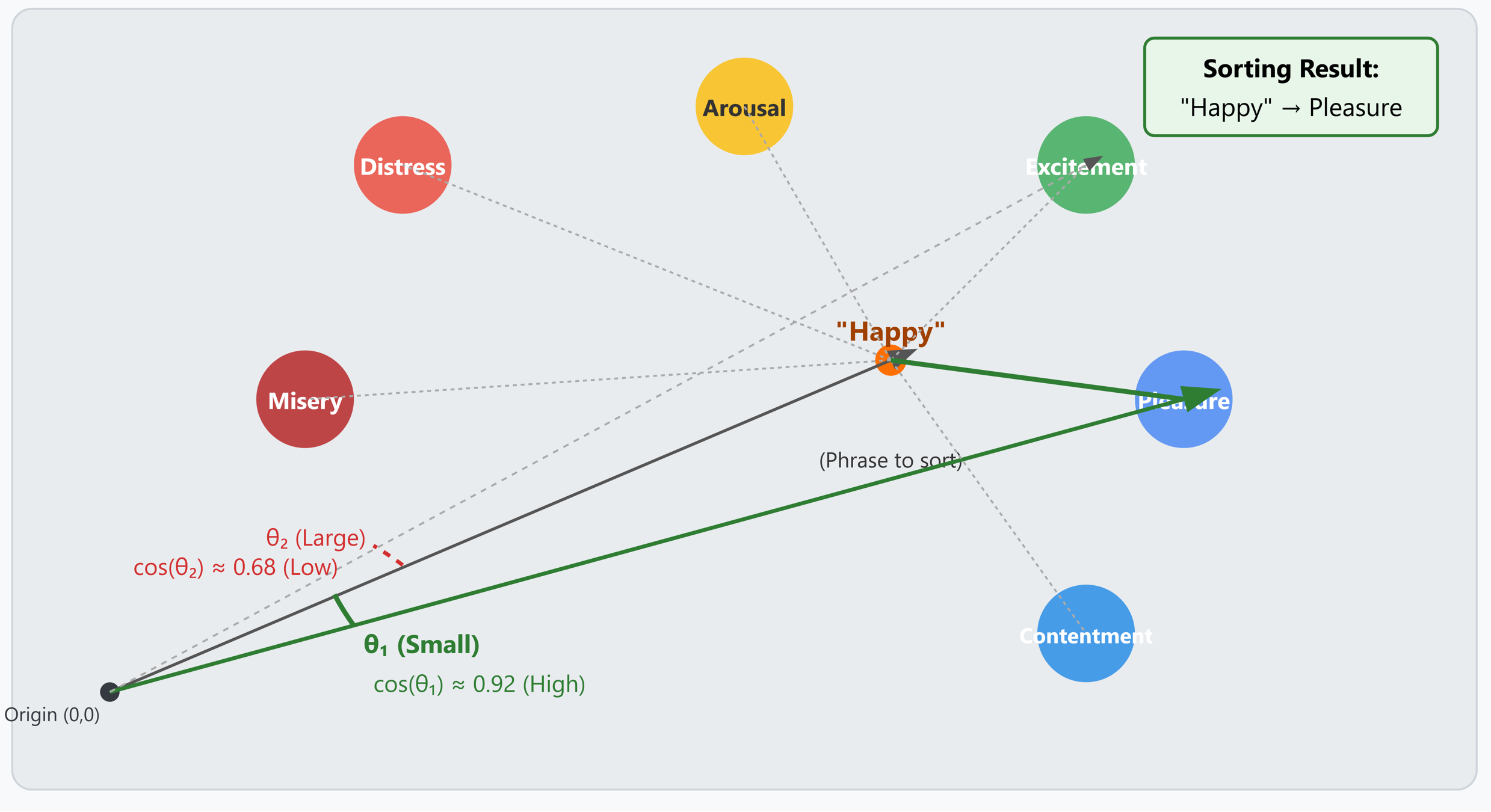}
  \caption{Cosine Similarity Sorting in Embedding Space (Conceptual 2D Projection of High-Dimensional Space)}
  \label{fig:cosine}
\end{figure}

At inference, for each fine-grained term $t$, we have $M{=}50$ phrases $\{t_m\}_{m=1}^{50}$. Each sentence is embedded $e(t_m)$ and assigned to the primary emotion whose centroid has the highest cosine similarity with ( this mechanism is illustrated in Figure \ref{fig:cosine}) :

\begin{equation}
\hat{c}_m \;=\; \arg\max_{c}\; \frac{e(t_m)\cdot \mu_c}{\|e(t_m)\|_2\,\|\mu_c\|_2}.
\end{equation}

After assigning all fine-grained emotion phrases to the primary emotion categories, we form the \emph{assignment distribution} $p{(t)}$ for each of the fine-grained terms, equivalent to the frequency of placement of 28 words into eight categories by Russell (see Table \ref{tab:freq_28_words_compact})

\[
\begin{aligned}
p^{(t)} \;=\; \big[\, p_{\text{Pleased}}(t),\; p_{\text{Excited}}(t),\; \ldots,\; p_{\text{Contented}}(t) \,\big],\\
\sum_{c\in\mathcal{C}} p_c(t) = 1.
\end{aligned}
\]

\subsubsection{Polar Geometry and Metrics}
\label{sec:polar}
To obtain comparable results to Russell's experiment, we employ the same approach, which provides a measure of central tendency and a measure of precision in terms of the vectorial angle and scalar magnitude \cite{ross1938circular}. This allows us to convert each fine-term’s assignment distribution (from the fine to primary) into a position on the valence-arousal circle with polar angle and precision (see Figure \ref{fig:28emotion}).
To do so, we place primary emotions at fixed angles on the unit circle:
\[
\Theta=\{0^\circ,45^\circ,90^\circ,135^\circ,180^\circ,225^\circ,270^\circ,315^\circ\}
\quad
\]
\[
\begin{aligned}
\text{for}\quad
\{\textit{Pleased},\textit{Excited},\textit{Aroused},\textit{Distressed},\textit{Miserable},\\
\textit{Depressed},\textit{Sleepy},\textit{Contented}\}.
\end{aligned}
\]
For a fine-grained emotion term $t$ with an assignment distribution $p^{(t)}$, we compute
\[
\begin{aligned}
x_t&=\sum_{c} p^{(t)}_c\cos\theta_c, \qquad
y_t=\sum_{c} p^{(t)}_c\sin\theta_c,\\
\theta_t&=\operatorname{atan2}(y_t,x_t), \qquad
P_t=\sqrt{x_t^2+y_t^2}\in[0,1].
\end{aligned}
\]
where 

\[
p^{(t)}_c \;=\; \frac{1}{M}\sum_{m=1}^{M}\text{1}\{\hat{c}_m=c\}, \qquad \sum_{c\in\mathcal{C}} p^{(t)}_c = 1.
\]

and $\theta_t$ is the polar \emph{angle} (placement where the term sits in valence–arousal space) and $P_t$ the \emph{resultant precision} (concentration):

\begin{itemize}
    \item P near 1 → the distribution strongly favors one primary emotion (high agreement/clarity).
    \item P near 0 → the distribution is spread across primary emotion categories (ambiguous/mixed)
\end{itemize}

We use these polar angles and precision to do a circumplex mapping and project each of the fine-grained terms' assignment distribution \( p(t) \) into the Valence-Arousal space.

\begin{figure}[h]
  \centering
  \includegraphics[width=\linewidth]{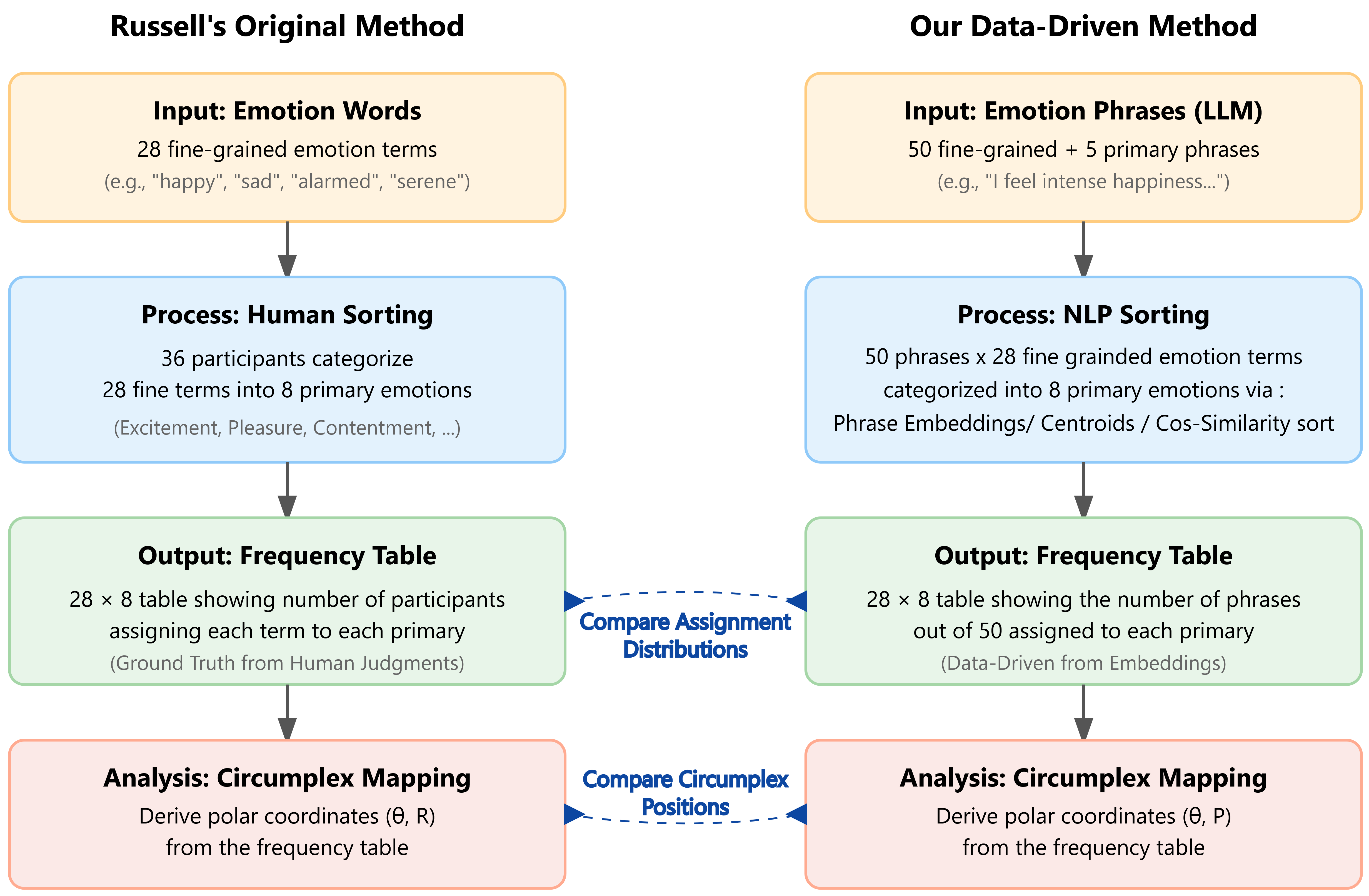}
  \caption{Method Comparison: Human Sorting vs. Embedding Similarity}
  \label{fig:compare}
\end{figure}

Figure \ref{fig:compare} positions our method alongside Russell's original approach, emphasizing the structural parallel. Russell presented 36 participants with emotion words and asked them to sort these words into eight primary emotion groups, producing an agreement/frequency table that reflected the number of participants assigning each fine-grained term to each primary category. This ground-truth structure is derived from collective human judgment. Our method replaces human participants with 50 LLM-generated phrases per emotion term, replaces human judgment with NLP-based sorting, and produces a similar frequency table reflecting the count of phrases assigned to each primary category. The key innovation is that we compare structures by converting both tables into polar coordinates and examining angular positions and precision values on the circumplex.

The comparison occurs at two levels, as shown in Figure \ref{fig:compare}. First, we compare the tables themselves: does our cosine-similarity sorting produce assignment distributions similar to the human agreement patterns Russell observed? Second, we compare the resulting circumplex positions: when both tables are converted to polar coordinates, do fine-grained terms land in similar angular regions with comparable precision? This dual validation ensures our method captures both the discrete assignment patterns (which primaries attract which fine-grained terms) and the continuous geometric structure (where terms sit in valence-arousal space).

This framework allows us to assess whether Transformer embeddings encode a psychologically grounded structure of emotion.\\

\subsection{Results of Framework 2}
\vspace{-0.6\baselineskip}
\begin{table}[htbp]
\centering
\caption{Frequency of Placement of 28 Words Into Eight Categories (Our results)}
\label{tab:fine_vs_primary_counts_no_pca}
\begingroup
\scriptsize
\setlength{\tabcolsep}{2pt}
\renewcommand{\arraystretch}{0.82}
\begin{tabular}{@{}l*{8}{c}@{}}
\toprule
\multicolumn{1}{c}{\makecell[c]{Fine-grained\\Emotion}} & \multicolumn{8}{c}{Predicted Primary Emotion} \\
\cmidrule(lr){2-9}
 & Pleasure & Excitement & Arousal & Distress & Misery & Depression & \makecell{Sleepi-\\ness} & \makecell{Content-\\ment}\\
\midrule
Happy      & 8  & 22 &    & 2  & 3  & 4  & 3  & 8  \\
Delighted  & 26 & 22 &    &    &    &    & 1  & 1  \\
Excited    & 10 & 38 &    &    &    &    & 2  &    \\
Astonished & 4  & 32 &    & 2  & 5  & 1  &    & 6  \\
Aroused    &    & 37 & 1  & 4  & 1  & 2  &    & 5  \\
Tense      &    & 24 & 3  & 10 &    & 5  & 8  &    \\
Alarmed    &    & 11 &    & 34 & 1  & 2  & 2  &    \\
Angry      & 1  & 9  &    & 29 & 3  & 3  & 5  &    \\
Afraid     &    & 9  & 1  & 27 & 3  & 3  & 4  & 3  \\
Annoyed    &    & 2  &    & 20 &    & 15 & 12 & 1  \\
Distressed & 1  & 8  &    & 36 & 1  & 3  & 1  &    \\
Frustrated & 1  & 8  &    & 28 &    & 10 & 2  & 1  \\
Miserable  &    & 10 &    & 9  & 11 & 15 & 5  &    \\
Sad        & 3  & 11 &    & 14 & 9  & 10 & 2  & 1  \\
Gloomy     & 2  & 8  &    & 1  & 6  & 31 &    & 2  \\
Depressed  & 1  &    & 3  & 22 & 4  & 19 & 1  &    \\
Bored      &    & 5  &    & 8  & 2  & 29 & 5  & 1  \\
Droopy     &    & 2  & 2  & 3  &    & 32 & 11 &    \\
Tired      & 1  & 5  & 3  & 7  & 2  & 8  & 23 & 1  \\
Sleepy     &    & 9  & 1  & 2  &    & 15 & 22 & 1  \\
Calm       & 2  & 10 & 1  & 12 &    & 5  & 2  & 18 \\
Relaxed    &    & 4  & 2  & 5  &    & 10 & 1  & 28 \\
Satisfied  & 23 & 4  &    & 11 & 2  &    &    & 10 \\
At ease    & 1  & 5  &    &    &    &    &    & 44 \\
Content    & 7  & 2  & 5  & 2  & 2  & 8  & 2  & 22 \\
Serene     &    & 1  & 1  & 1  & 1  & 4  &    & 42 \\
Glad       & 20 & 16 &    & 5  &    & 1  & 1  & 7  \\
Pleased    & 41 & 7  &    &    &    &    &    & 2  \\
\bottomrule
\end{tabular}
\endgroup
\end{table}

Table \ref{tab:fine_vs_primary_counts_no_pca} presents the full prediction matrix showing how our cosine-similarity sorting assigned each of the 28 fine-grained emotion terms to the eight primary categories. The matrix reveals several patterns consistent with Russell's original findings:
Our approach achieves strong agreement for emotionally unambiguous terms. For instance, \textbf{Pleased} was assigned to the Pleasure category 41 out of 50 times (82\%), while \textbf{Excited} mapped to the Excitement category 38 times (76\%). Terms with clearer valence-arousal signatures exhibited higher concentration (precision P).

More ambiguous terms showed distributed assignments reflecting genuine emotional complexity. \textbf{Annoyed} scattered across Distress (20; 40\%), Depression (15; 30\%), and Sleepiness (12; 24\%), mirroring the term's position near the boundary of multiple primary emotions. Similarly, \textbf{Tired} was distributed across Sleepiness (23; 46\%), Contentment (8; 16\%), and Depression (7; 14\%), capturing its low-arousal but valence-ambiguous nature.

\begin{figure}[htpb]
  \centering
  \includegraphics[width=0.9\linewidth]{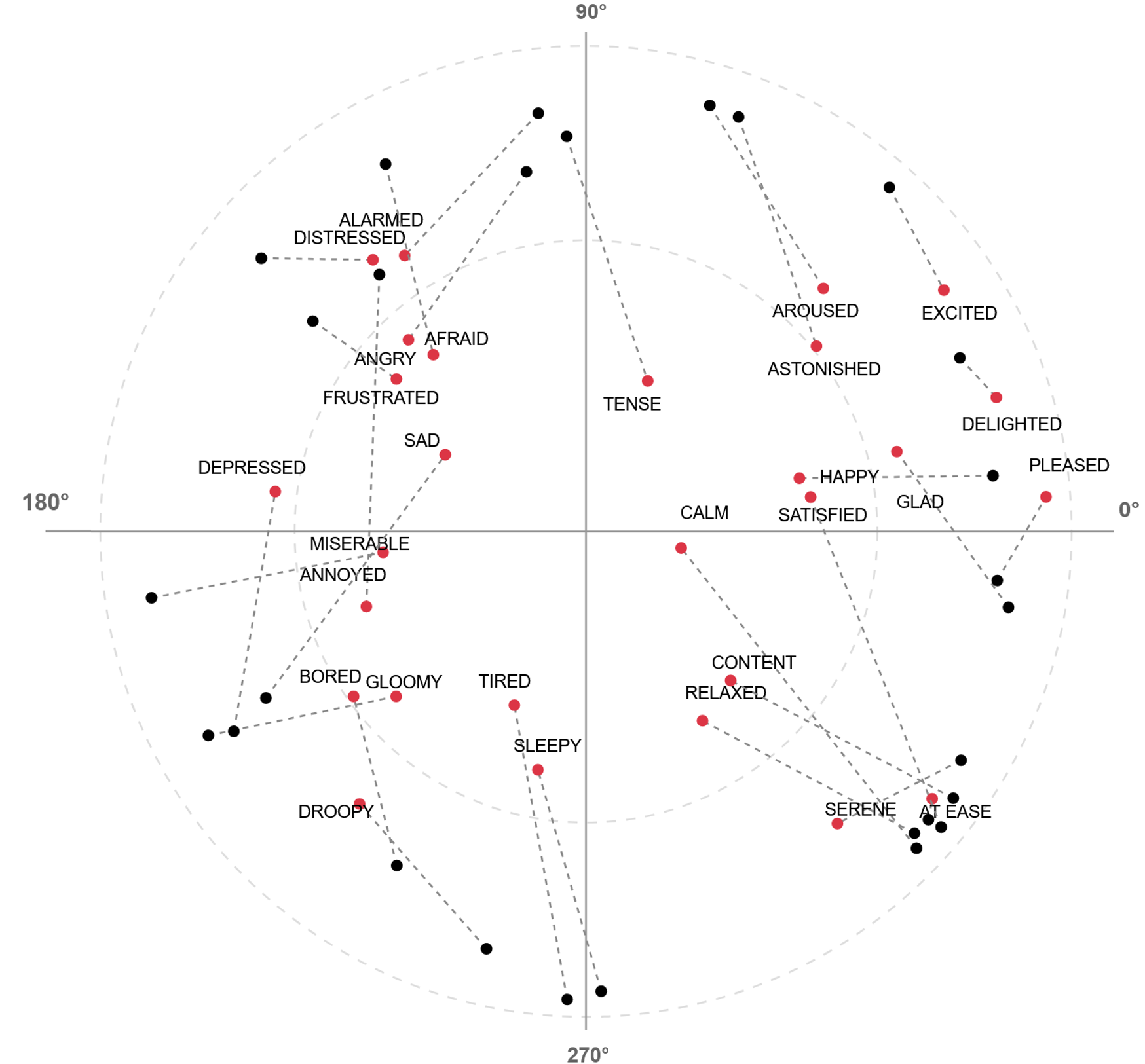}
  \caption{Comparison: Russell's Original Positions (1980) in black vs Positions based on our results in red}
  \label{fig:our_result_vs_russel}
\end{figure}

Converting the prediction distributions to polar coordinates revealed close alignment with Russell's reference angles for most terms. Figure \ref{fig:our_result_vs_russel} presents our data-driven results, showing the computed polar positions (\(\theta_t\), \(P_t\)) for each of the 28 fine-grained emotions based on cosine-similarity sorting of 50 LLM-generated phrases per term. The Figure also overlays our results (red dots) with Russell's original human-participant placements (black dots), with dashed lines connecting corresponding terms to visualize the position deviations.

Our results demonstrate strong overall fidelity to Russell's circumplex structure. High-valence terms (Happy, Delighted, Pleased, Glad) cluster tightly in the 0°-45° quadrant. Figure \ref{fig:our_result_vs_russel} shows minimal displacement between red and black dots in this region, indicating our embedding-based approach accurately captures the positive-arousal sector that Russell's participants consistently identified.

Negative high-arousal emotions (Angry, Afraid, Alarmed, Distressed) concentrate near 90°-135° in both mappings, with the dashed lines in Figure \ref{fig:our_result_vs_russel} revealing slightly longer connections in this quadrant. Notably, \textbf{Tense} deviates from its Russell positioning, which places it near the vertical arousal axis in the negative quadrant, whereas our embeddings place it at lower arousal with slightly positive valence.

Discrepancies occur in regions of low arousal. In the depressive sector (180°-270°), our approach brings \textbf{Bored}, \textbf{Gloomy}, \textbf{Depressed}, and \textbf{Droopy} into a tighter cluster than Russell's more dispersed arrangement. Figure \ref{fig:our_result_vs_russel} shows longer dashed lines here as well, particularly for \textbf{Sad}, which in our results shifts toward less negative valence and higher arousal compared to Russell's placement in the low-arousal depressive region, suggesting that the embedding space captures \textbf{Sad} in more activated, expressive contexts rather than as a purely low-energy negative state.

In the contentment sector (270°-0°), terms like \textbf{Relaxed}, \textbf{Content}, \textbf{Serene}, and \textbf{At ease} maintain appropriate positions in the low-arousal positive quadrant in our results. However, Figure \ref{fig:our_result_vs_russel} shows that \textbf{Calm} and \textbf{Satisfied} deviate substantially, shifting toward the horizontal positive valence axis near Pleasure. This shift suggests the embedding space emphasizes their positive semantic content over their low-arousal character.

Terms exhibiting the closest correspondence \textbf{Excited}, \textbf{At ease},\textbf{Pleased}, \textbf{Serene}, and \textbf{Distressed} show neighbouring red and black dots in Figure \ref{fig:our_result_vs_russel}, demonstrating emotionally unambiguous character.

Precision values validate the circumplex topology: terms with high precision appear near primary emotion centers in Figure \ref{fig:our_result_vs_russel}, while low-precision terms occupy boundary regions where Russell's participants also showed disagreement. Critically, no term violates the circular structure; even ambiguous terms' placements respect neighborhood relations, with confusions occurring between adjacent rather than opposite primaries.

\subsubsection{Angular Alignment Analysis
} 
\begin{figure}[hpb!]
  \centering
  \includegraphics[width=\linewidth]{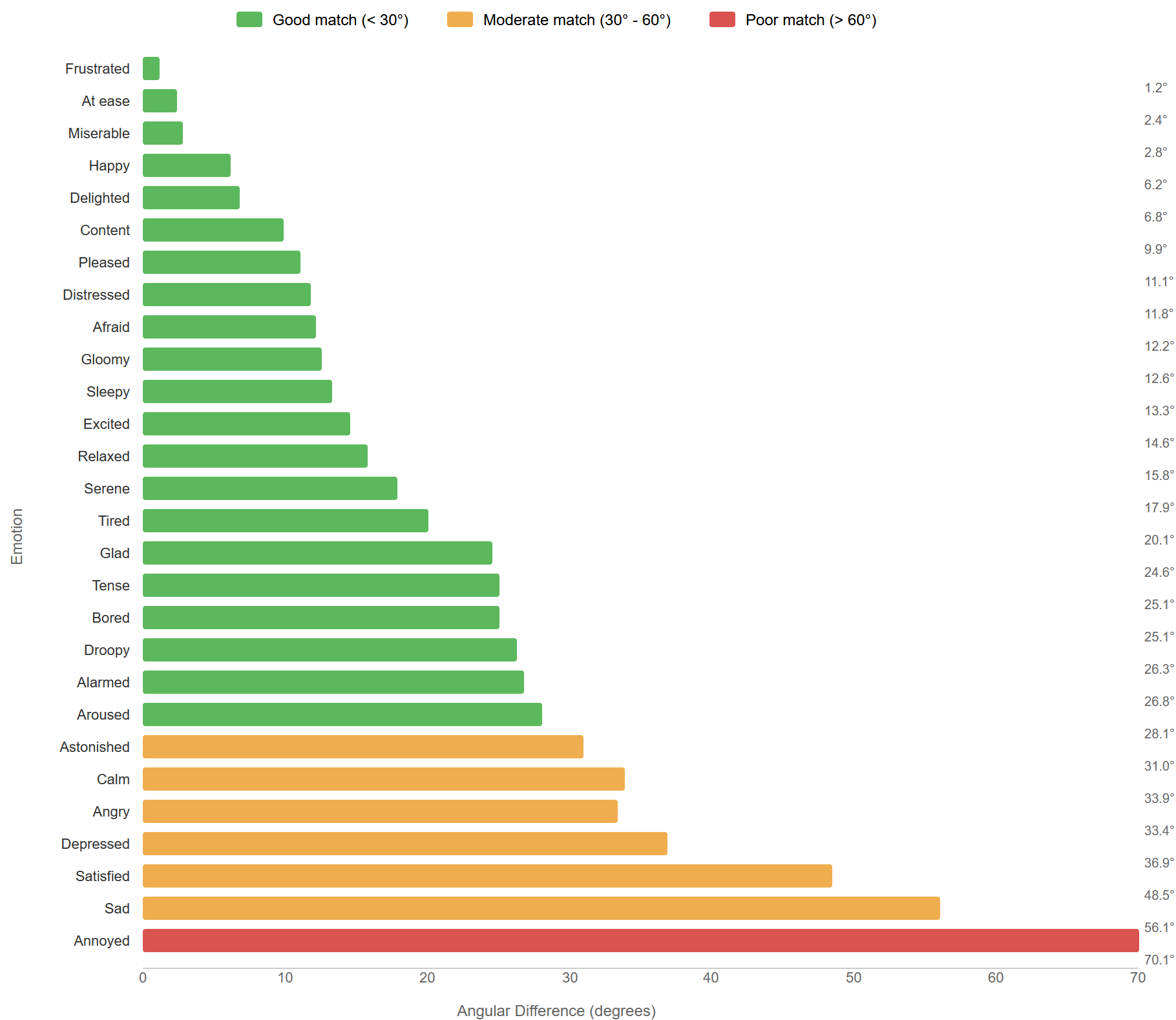}
  \caption{Angular Difference Between Russell's Positions and Our Positions }
  \label{fig:Angular}
\end{figure}

In Figure \ref{fig:Angular}, we quantify the angular deviation between our embedding-based placements and Russell's reference coordinates for all 28 terms. By applying \textbf{Angular error} if a reference angle $\theta^{\ast}_t$ is specified (e.g., from Russell’s mapping), we report
\[
\Delta\theta_t = \min\big(|\theta_t-\theta^{\ast}_t|,\, 360^\circ-|\theta_t-\theta^{\ast}_t|\big)\, .
\]
\vspace{-0.6\baselineskip}

The distribution reveals strong overall alignment: 23 out of 28 terms (82\%) achieve good-to-moderate match quality ($\Delta\theta < 60^{\circ}$), with 21 terms (75\%) falling into the "good match" category ($\Delta\theta < 30^{\circ}$).

The best-performing terms demonstrate near-perfect correspondence: \textbf{Frustrated} ($\Delta\theta$ = 1.2°), \textbf{At ease} ($\Delta\theta$ = 2.4°), \textbf{Miserable} ($\Delta\theta$ = 2.8°), \textbf{Happy} ($\Delta\theta$ = 6.2°), \textbf{Delighted} ($\Delta\theta$ = 6.8°), \textbf{Content} ($\Delta\theta$ = 9.9°), \textbf{Pleased} ($\Delta\theta$ = 9.9°), and \textbf{Distressed} ($\Delta\theta$ = 11.1°). These terms span all quadrants of the circumplex, indicating our method successfully recovers both positive/negative valence and high/low arousal distinctions.

Mid-range deviations (15°-30°) appear for terms in transitional regions: \textbf{Aroused} ($\Delta\theta$ = 26.8°), \textbf{Alarmed} ($\Delta\theta$ = 26.3°), \textbf{Droopy} ($\Delta\theta$ = 25.1°), and \textbf{Bored} ($\Delta\theta$ = 25.1°). These moderate misalignments reflect genuine ambiguity.

Moderate-match terms (30°-60°) include \textbf{Astonished} ($\Delta\theta$ = 28.1°), \textbf{Calm} ($\Delta\theta$ = 31.0°), \textbf{Angry} ($\Delta\theta$ = 33.9°), \textbf{Depressed} ($\Delta\theta$ = 33.4°), \textbf{Satisfied} ($\Delta\theta$ = 36.9°), and \textbf{Sad} ($\Delta\theta$ = 48.5°). The deviation or displacement of these fine-grained terms is explained by their assignment distribution, suggesting that their valence and arousal aspects vary in the embedding space. 

Only one term exhibits poor alignment: \textbf{Annoyed} ($\Delta\theta$ = 70.1°). This outlier reflects the term's semantic complexity; it can denote mild irritation (low arousal, slightly negative), frustration (moderate arousal, negative), or anger precursor (higher arousal, strongly negative). The embedding space appears to emphasize the low-arousal interpretation, placing it closer to Misery, whereas in Russell's results, it is positioned closer to Distress.

\subsubsection{Precision Comparison} 
\begin{figure}[h]
  \centering
  \includegraphics[width=\linewidth]{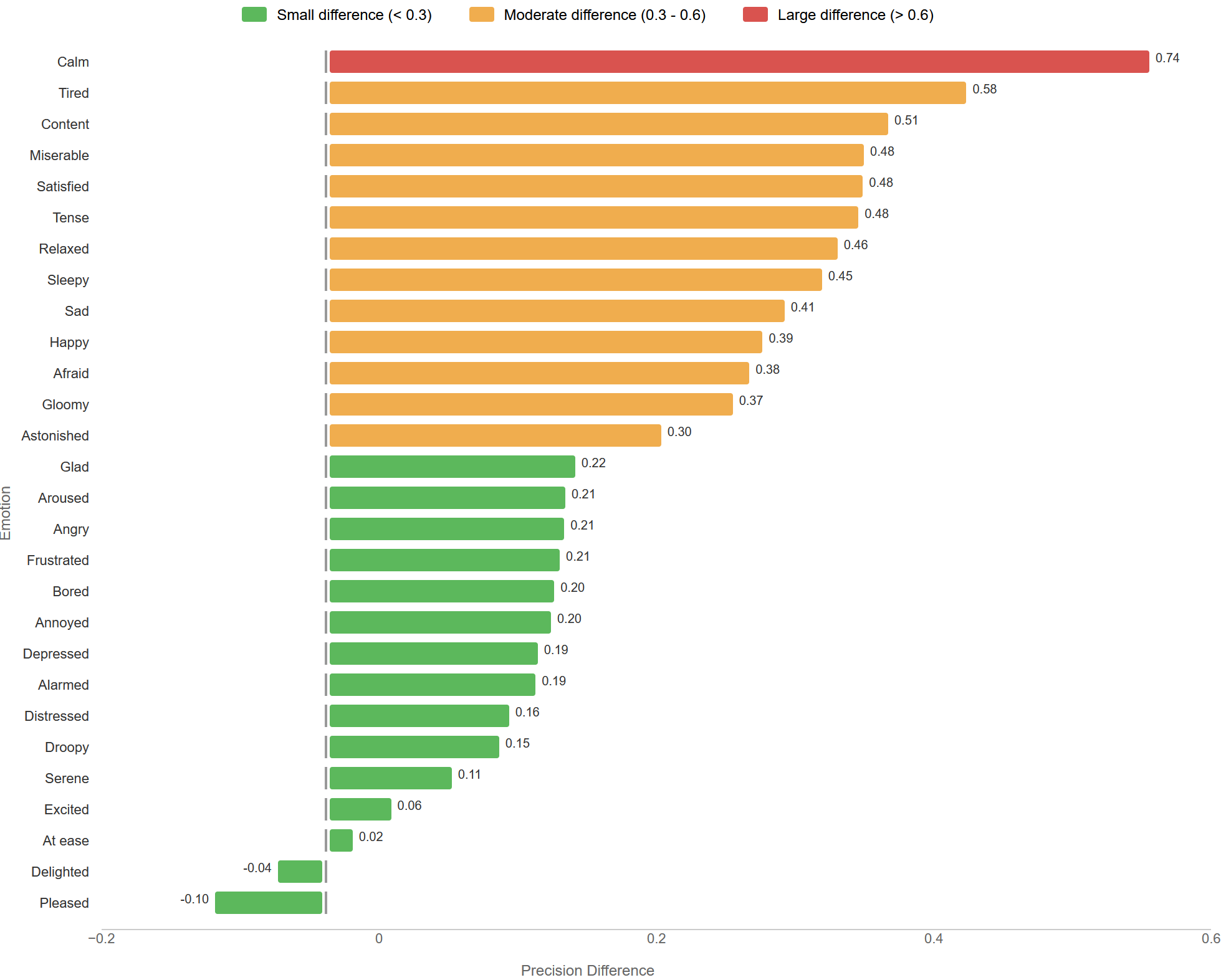}
  \caption{Precision Difference Between Russell's Positions and Our Positions }
  \label{fig:Precision}
\end{figure}

In Figure \ref{fig:Precision} we examine precision differences. 
\[
\Delta P = \ P_{\text{ours}} - P_{\text{Russell}} \ .
\]
to reveal how embedding-based concentration compares to human agreement patterns. Negative values indicate our approach produced higher precision than Russell's human data, while positive values show lower precision.

Terms with near-zero precision difference ($|\Delta P| < 0.2$) demonstrate closest correspondence to human sorting consistency: \textbf{At ease} ($\Delta P$ = 0.02), \textbf{Excited} ($\Delta P$ = 0.06), \textbf{Serene} ($\Delta P$ = 0.11), \textbf{Droopy} ($\Delta P$ = 0.15), \textbf{Distressed} ($\Delta P$ = 0.16), \textbf{Alarmed} ($\Delta P$ = 0.19), \textbf{Depressed} ($\Delta P$ = 0.19), \textbf{Annoyed} ($\Delta P$ = 0.20), and \textbf{Bored} ($\Delta P$ = 0.20). These terms span all circumplex sectors, indicating our method captures human-like disambiguation patterns across the affective space.

Small negative values for \textbf{Pleased} ($\Delta P$ = - 0.10) and \textbf{Delighted} ($\Delta P$ = - 0.04) indicate our embeddings produce slightly higher consensus than Russell's participants; all 50 phrases converge tightly to the Pleasure category.

Moderate precision differences ($0.3 < \Delta P < 0.6$) appear for boundary terms: \textbf{Astonished} ($\Delta P$ = 0.30), \textbf{Gloomy} ($\Delta P$ = 0.37), \textbf{Afraid} ($\Delta P$ = 0.38), \textbf{Happy} ($\Delta P$ = 0.39), \textbf{Sad} ($\Delta P$ = 0.41), \textbf{Sleepy} ($\Delta P$ = 0.45), \textbf{Relaxed} ($\Delta P$ = 0.46), \textbf{Tense} ($\Delta P$ = 0.48), \textbf{Miserable} ($\Delta P$ = 0.48), \textbf{Satisfied} ($\Delta P$ = 0.48), \textbf{Content} ($\Delta P$ = 0.51), and \textbf{Tired} ($\Delta P$ = 0.58). Lower precision in our approach reflects the embedding space spreading these terms across multiple primaries, suggesting the semantic neighborhoods overlap more in the latent space than in human psychology.

The largest precision difference occurs for \textbf{Calm} ($\Delta P$ = 0.74). Russell's participants showed strong consensus ($P_{\text{Russell}} \approx 0.82$), placing it in Contentment (see Table \ref{tab:freq_28_words_compact}), whereas our phrases scatter across Contentment (12/50), Pleasure (10/50), and Sleepiness (18/50), yielding P = 0.36. This dispersion indicates \textbf{Calm} phrases in the embedding space activate multiple low-arousal concepts that humans more clearly distinguish.

Our results demonstrate that embedding-based sorting recovers Russell's circumplex geometry with moderate to strong angular fidelity and interpretable precision patterns.

\section{Discussion \& Limitations}
\label{sec:four}

The two studies we conducted to reproduce the circular ordering (Task 1) and category sorting (Task 2) provide complementary insights into how Transformer embeddings encode affective structure. 

In our first study, we investigate alignments between the learned emotional representation and Russell’s psychological circumplex model of affect through a data-driven deep learning approach. The results show the importance of deep learning models, dataset characteristics, and fused modalities in shaping the representations learned. In contrast to generic models, fine-tuned models led to more plausible emotional representations due to the complex nature of emotions. The only case where the generic model yielded similar results was with the dataset extracted from CoLiTec corpora. Since the dataset explicitly includes affective words, it led to a clear separation of emotional clusters.

The more naturalistic MSP-Podcast dataset presented a more significant challenge due to its inherent complexity and variability. Combining both modalities has shown to be more compelling in finding the alignment with Russell's model; this implies that modality fusion allows for a richer and more nuanced representation space. However, there are limitations to our study. First, the use of fine-tuned models and their inherent biases. Second, the dimensionality reduction and the possible oversimplification of the emotional space. Lastly, the datasets used do not cover the entire spectrum of human emotion. Going forward, we will explore fully unsupervised learning models, such as autoencoders, to discover emotional representations without prior knowledge of emotions. 

In our second study, we address the fine-tuning dependency identified in Task 1 by investigating whether a generic, pre-trained Transformer model without any explicit emotion-specific training can reproduce Russell's category sorting results. Unlike Task 1, where fine-tuned models demonstrated clear advantages on naturalistic datasets, in Task 2, we explore the zero-shot affective geometry encoded in Transformer embeddings trained purely on language modeling objectives.

We faced methodological constraints in replicating Task 2 in a data-driven manner; no existing public dataset contains the specific 28 terms Russell used with sufficient phrase diversity to support our embedding-based sorting approach. This scarcity necessitated the generation of synthetic phrases via large language models, introducing a controlled artificial data source.


A single emotion word, as used in Russell's study, is an abstraction; real emotional expression in language is heterogeneous and context-dependent. Our approach operates at a higher level of semantic complexity: each fine-grained emotion is represented by 50 distinct phrases that express the emotion in varied linguistic contexts (e.g., for \textbf{Happy}: "I feel happy about the outcome," "She expressed happiness through her smile," "The news made him incredibly happy"). This forms a distributed cloud in semantic space, and our centroid-based sorting must aggregate across this diversity.

Despite using a base RoBERTa model without emotion finetuning, in Task 2, we achieved remarkable alignment with Russell's circumplex. Three-quarters of the 28 fine-grained terms (75\%) landed within 30° of their reference positions. This result is noteworthy, as it demonstrates that dimensional affect structure is not solely an artifact of supervised emotion training, but is instead encoded in the semantic relationships learned from general language patterns. 

Moderate precision differences at category boundaries indicate greater semantic variability, which the embedding model detects. For example, \textbf{Calm} phrases span Contentment, Sleepiness, and Pleasure, showing that calmness includes diverse expressions (e.g., peaceful, relaxed, serene). In contrast, terms like Pleased (P = 0.82), Excited (P = 0.76), and Distressed (P = 0.72) exhibited high precision. This implies that these emotions are semantically consistent: different linguistic descriptions do not change their core affective meaning, enabling our approach to sort them with high confidence.
\vspace{-0.6\baselineskip}
\begin{table}[htpb]
\centering
\caption{Summary: Task~1 (Circular Ordering) vs.\ Task~2 (Category Sort)}
\label{tab:task_comparison}
\scriptsize
\begin{tabular}{lll}
\hline
 & \textbf{Task~1} & \textbf{Task~2} \\
\hline
Modalities & Text, audio, fusion & Text only \\
Models & Fine-tuned RoBERTa \& wav2vec & Generic RoBERTa-base \\
Data & CoLiTec, TESS, MSP-Podcast & LLM-generated phrases \\
Metric & Cyclic mismatch count & $\Delta\theta$, $\Delta P$ \\
Best result & 0 mismatches (fusion) & 75\% terms $\Delta\theta < 30^{\circ}$ \\
\hline
\end{tabular}
\end{table}

These constraints, text-only modality, a generic model, and LLM-generated phrases, create a complementary experimental setup. Where Task 1 explored multimodal fusion and fine-tuned representations on naturalistic speech, Task 2 isolates the lexical-semantic structure in general-purpose embeddings using controlled linguistic stimuli. Together, they frame the question: \textit{ Does the circumplex emerge from learned affect-specific patterns (Task 1) or from fundamental semantic organization present even without emotion training (Task 2)?}

Our findings are robust within the circumplex framework, but should be interpreted as diagnostic rather than prescriptive. Our validation that Transformer-based model embeddings recover the circumplex does not imply that the circumplex is the \textbf{true} emotion model, but rather that it provides a psychologically grounded reference structure.

\section{Conclusion}

In this paper and through our work, we bridge classic psychological experiments with modern representation-space analyses through two complementary studies. We show that it is possible to reproduce both Russell's circular ordering task (Task 1) and category sorting task (Task 2) using novel, automated, data-driven approaches.

In Task 1, we leveraged Transformer-based models for feature extraction and dimensionality reduction, uncovering emotional manifolds. Through cosine similarity optimization over permutations, we recovered the circular order of eight primary emotions across text, audio, and fused modalities. Our results highlight that the success of this reproduction is fundamentally tied to the fusion of textual and audio modalities. Such fusion provides a richer and more nuanced representation of emotional states, achieving perfect alignment (zero cyclic mismatches).

In Task 2, we extended the investigation to fine-grained emotion semantics using a generic, pre-trained RoBERTa model without emotion-specific training. By introducing a cosine-similarity sorting mechanism that maps LLM-generated phrases per emotion term to a primary emotion centroid, we reproduced Russell's human category sort task patterns with 75\% of terms achieving good angular correspondence ($\Delta\theta< 30^{\circ}$). Notably, this success with generic embeddings indicates that dimensional affect structure arises from general language statistics. However, precision analysis exposed limitations: boundary terms and low-arousal states showed lower consensus than human participants, reflecting the increased semantic complexity of phrase-level representation compared to Russell's word-level sorting task.

Together, these studies provide convergent evidence that Transformer embeddings encode a topology consistent with psychological theories of emotion, but with important nuances. Task 1 demonstrates that multimodal fusion is essential for disambiguating naturalistic affective expressions, where context and prosody play a crucial role. Task 2 reveals that semantic structure supporting the circumplex exists even in unsupervised text representations; however, full human-like precision requires either emotion-specific adaptation or multimodal grounding. As in sensory systems where structure is inferred from systematic response patterns, affective structure can be probed through the geometry of learned representations. We do not claim a shared neural mechanism; rather, the analogy emphasizes that the most informative signal is the topological regularity itself.

Our future research will focus on overcoming the limitations detailed in Section~\ref{sec:four} by: employing our approach with fully unsupervised learning models, such as auto-encoders, expanding to more datasets to include more diverse emotional expressions, and exploring additional modalities along with evaluation metrics to achieve a more comprehensive analysis and further enhance the robustness of our findings.
 
The insights gained from our studies contribute to the current research on emotion representation and explainability in deep learning. By providing automated, scalable methods to validate a psychology-based emotion model against learned representations, our findings underscore the potential of multimodal techniques in capturing the complexity of human emotions while highlighting the value of examining both supervised and unsupervised latent spaces.

\section{Ethical Impact Statement}
As we move forward to explore and study the nuances of emotions, ethical considerations should be taken into account. Decoding emotions can be used responsibly in mental health therapy, care applications, or to enhance human-computer interaction. However, potential misuse can be in surveillance systems or manipulative marketing strategies. Developing and deploying emotion recognition technologies must be done responsibly, with transparency and respect for individual privacy.

In this context, our future work aligns with applications that are fully compliant with the European Union AI Act \cite{eu2024aiact}, ensuring that these technologies meet high standards of safety and fundamental rights.

%

\vfill

\end{document}